\pdfoutput=1

\documentclass[10pt,twocolumn,letterpaper]{article}

\usepackage[pagenumbers]{cvpr}

\usepackage{microtype}

\definecolor{cvprblue}{rgb}{0.21,0.49,0.74}
\usepackage[breaklinks,colorlinks,allcolors=cvprblue]{hyperref}
\usepackage{adjustbox}

\title{ClothCast: Generative Cloth World Modeling under Partial Observability}

\author{Yushen Hu\quad Yilong Song\textsuperscript{*}\quad Zhen Zhou\textsuperscript{*}\\
Apurv Kushwaha\quad Harshvardhan Chandirasekar\quad Chen Zhu-Tian\\
University of Minnesota, Department of Computer Science $\&$ Engineering\\
{\tt\small hu000809@umn.edu}\\
{\small \textsuperscript{*}Equal contribution}
}

\begin{document}
\maketitle
\begin{abstract}
World models support robotic manipulation by predicting how the scene evolves under a planned action. For garments this prediction is notoriously difficult: deformations are complex, and self-occlusion leaves much of the cloth unobserved. Planned gripper trajectories provide a useful motion prior because cloth held by a gripper moves with it, while the remaining deformation must be predicted. We introduce ClothCast, a world model that applies this prior to the cloth's points and learns the residual motion. ClothCast takes one RGB-D observation, the camera pose, the robot's own model and the planned gripper trajectories, with no cloth mesh or material parameters. It lifts the observation to a layered point set that includes hypothesized occluded layers, carries the points with a blended rigid prior derived from the gripper poses, and predicts the residual motion with a video-diffusion point-flow model. Over entire 4- to 11-second episodes at 10\,Hz, the moved points are rendered into depth, from which a geometry-conditioned completion model synthesizes RGB; after release, a hand-designed settling step replaces the learned motion. We further release ClothCastBench, $10{,}000$ simulated dual-arm cloth-manipulation episodes with the ground-truth cloth mesh, including occluded layers, at every frame. On 492 held-out episodes, ClothCast achieves 2.5$\times$ lower sequence-averaged visible-surface Chamfer distance than Flex-$\pi$, the strongest baseline in geometry (14.0\,mm vs.\ 35.6\,mm), and 3.7\,dB higher cloth-region PSNR than Cosmos-Predict2.5, the strongest baseline in cloth PSNR. On a 50-episode ablation subset, the rigid prior alone, applied to ClothCast's learned initial geometry with the same post-processing, already outperforms both geometry-capable baselines, and the learned residual lowers its error by about a third.
\end{abstract}

\section{Introduction}
\label{sec:intro}

\begin{figure*}[t]
  \centering
  \includegraphics[width=\textwidth]{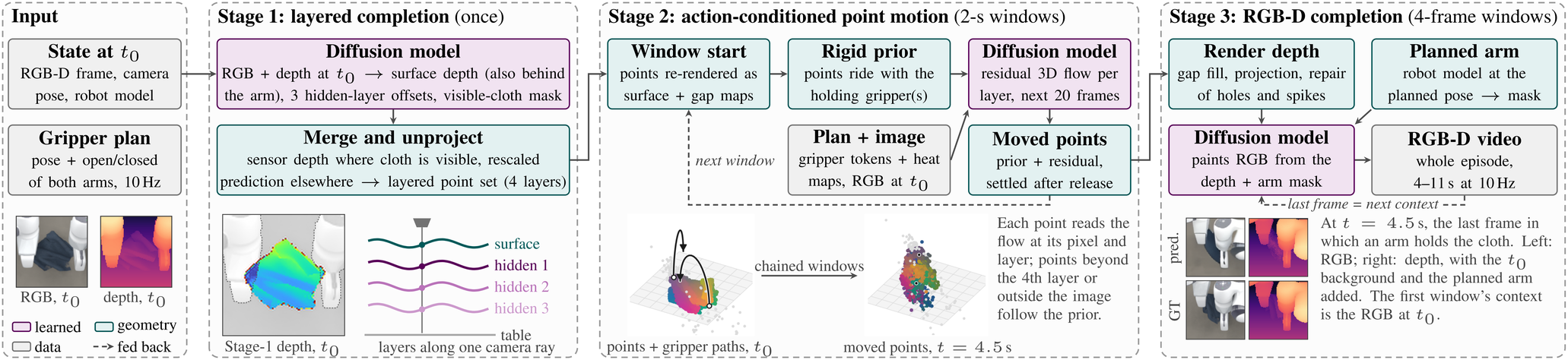}
  \caption{\textbf{ClothCast on a test episode.} Points are colored by the material coordinates of the nearest ground-truth cloth point at $t_0$, for visualization only (gray: none within 3\,cm); images are cropped around the cloth.}
  \label{fig:pipeline}
\end{figure*}

World models let a robot evaluate a planned action before executing it, by predicting how the scene will evolve. For garment manipulation this prediction is hard: a garment has no single pose but a wide range of folds, wrinkles, and layered self-contact~\cite{sanchez2018robotic,yin2021modeling}, and a single RGB-D image shows only part of it~\cite{longhini2024cloth}. We study forecasting the RGB-D observations of a cloth item throughout a dual-arm manipulation episode lasting several seconds, from the state observable at the start (one RGB-D observation, the camera pose and the robot's own model) and the planned gripper trajectories, without a supplied cloth mesh or material parameters.

Our starting observation is that the action is more than a control signal: it constrains where the grasped cloth will be. Cloth attached to a gripper moves with it, which gives a motion prior and leaves only the remaining deformation to be predicted. Geometry-capable world models such as Flex-$\pi$~\cite{flexpi} and DepthWorld~\cite{depthworld} receive the action as a vector without explicit alignment to image pixels or cloth points; others ground it in the camera view~\cite{jiang2025evac,su2025mtvworld} or as 3D robot point flows~\cite{huang2026pointworld}, but still learn the resulting scene motion. Many cloth dynamics models apply grasp motion to the cloth state directly, but need a category-specific reconstruction~\cite{huang2022medor} or, when physics-based, explicit geometry and material parameters~\cite{zheng2024diffcp}. We apply a blended rigid prior directly to an estimated cloth point set: on our 50-episode ablation subset, this prior, with ClothCast's learned initial geometry and the same post-processing but no learned motion, already gives a lower geometry error than both geometry-capable world models (21.2\,mm against 34.2 and 67.2\,mm; \cref{sec:exp_ablation}).

We build \textbf{ClothCast} on this observation. It represents the cloth as a set of 3D points and expresses the planned action as their motion: the blended rigid prior moves the points with the gripper poses, and a video-diffusion backbone built on the Ctrl-World architecture~\cite{guo2026ctrl} predicts only the residual, how the cloth departs from this prior, as dense flow fields in image-space canvases, with the gripper trajectories also rendered as spatial channels. Two further stages follow from this choice: because the dynamics stage moves a fixed point set, a first stage lifts the initial observation to a layered point set that includes hypothesized occluded layers, for cloth that may become visible later; because the prediction target is the RGB-D observation sequence, a completion stage renders the moved points into depth and synthesizes RGB from it, autoregressively from the initial observation and conditioned on the arm silhouette along the planned trajectory. Chaining 2-second windows covers entire 4--11-second episodes at 10\,Hz, with a hand-designed settling step after release (\cref{fig:pipeline}).

Training this model needs supervision that real RGB-D recordings cannot provide: the geometry of occluded cloth, and where every cloth point moves. We therefore release \textbf{ClothCastBench}, $10{,}000$ simulated episodes of scripted dual-arm manipulation of pants, long-sleeve shirts, t-shirts and square sheets with varied materials and textures, each with a $256\times256$ RGB-D sequence at 10\,Hz and the ground-truth cloth mesh at every frame, from which both kinds of label are derived, and a fixed split that reserves $516$ episodes for evaluation.

On the 492 held-out episodes with outputs from all baselines, ClothCast achieves a sequence-averaged visible-surface Chamfer distance of $14.0$\,mm, 2.5$\times$ lower than Flex-$\pi$, the strongest baseline in geometry, and raises cloth-region PSNR by 3.7\,dB over Cosmos-Predict2.5~\cite{cosmos}, the strongest baseline in cloth PSNR (\cref{sec:experiments}). On the ablation subset, adding the learned residual to the prior lowers the error from 21.2 to 14.5\,mm, with the largest gains while the cloth is being manipulated (\cref{sec:exp_ablation}); against the simulator's layered ground truth, the occluded layers lower the full-geometry Chamfer distance from 16.4 to 13.4\,mm (\cref{sec:exp_occluded}). On 84 real dual-arm test episodes, conditioned on the recorded executed gripper trajectories, the geometry stages, trained only in simulation, reach 35.3\,mm against the sensor depth at a coverage of 0.66, compared with 44.9\,mm for the prior without learned motion; no baseline was run on real data.
Our contributions are:
\begin{itemize}
    \item \textbf{Motion representation.} We formulate cloth motion as a learned residual over a blended rigid prior that applies the planned gripper poses directly to the cloth's points.
    \item \textbf{Model.} ClothCast, a point-flow world model that combines this motion representation with a video-diffusion backbone, starts from a layered point set with hypothesized occluded cloth, and renders RGB-D sequences over 4--11-second episodes, with lower geometry error than both geometry-capable world models at every reported horizon.
    \item \textbf{Benchmark.} ClothCastBench, $10{,}000$ dual-arm manipulation episodes with ground-truth geometry for all visible and occluded cloth at every frame, enabling supervision of full-geometry and point-motion predictions.
\end{itemize}

\section{Related Work}
\label{sec:related}

\begin{table*}[t]
  \centering
  \scriptsize
  \setlength{\tabcolsep}{4pt}
  \caption{\textbf{Positioning} of ClothCast. $^\dagger$Flex-$\pi$ natively generates actions; we condition it on the planned actions (\cref{sec:exp_setup}).}
  \label{tab:positioning}
  \begin{adjustbox}{max width=\textwidth}
  \begin{tabular}{@{}lllll@{}}
    \toprule
    Method & Scene state & How the action enters & Geometry of initially occluded cloth & Test-time inputs \\
    \midrule
    Ctrl-World, Cosmos, V-JEPA 2-AC & pixels, latents or features & action vector or tokens & not explicitly represented & image(s) (V-JEPA: + robot state) \\
    DepthWorld & per-pixel depth & embedded action vector & not explicitly represented & RGB-D \\
    Flex-$\pi$ & per-pixel pointmap & action vector$^\dagger$ & not explicitly represented & RGB-D, robot state, intrinsics \\
    EnerVerse-AC, MTV-World & pixels & image-aligned action maps & not explicitly represented & images, camera \\
    PointWorld & observed 3D points & robot 3D point flows & observed points only & RGB-D, camera, robot model \\
    \midrule
    VCD, PGND & observed particles (PGND: + grid) & constraints on grasped particles & observed points only & depth or sparse-view RGB-D \\
    MEDOR & reconstructed mesh & motion of grasped vertex & reconstructed, category-specific model & depth \\
    UniClothDiff & full cloth state & gripper actions & estimated from a template & point clouds, template, state history \\
    \midrule
    ClothCast & layered 3D points & rigid prior on cloth points, tokens, heat maps & hypothesized layers, one model & RGB-D, camera, robot model \\
    \bottomrule
  \end{tabular}
  \end{adjustbox}
\end{table*}

Video world models predict scenes from images, and cloth dynamics models from an explicit cloth state; \cref{tab:positioning} compares both on how the action enters and what each must know about the cloth at the start.

\paragraph{Action-conditioned world models.}
World models forecast the consequences of actions so that plans can be evaluated before execution~\cite{ebert2018visualforesight,hafner2019planet}, for fabric as RGB-D video in VisuoSpatial Foresight~\cite{hoque2020vsf}. Ctrl-World~\cite{guo2026ctrl} adds frame-level action conditioning to Stable Video Diffusion~\cite{blattmann2023stable}, Cosmos~\cite{cosmos} generates video, and V-JEPA~2-AC~\cite{vjepa2ac} predicts the features of a self-supervised video encoder, while DepthWorld~\cite{depthworld}, Flex-$\pi$~\cite{flexpi} and TesserAct~\cite{zhen2025tesseract} add depth, point maps or normals of the visible surface, aligned with the image.

\paragraph{How the action enters.}
These models take the action as a vector per frame or clip, as we also condition Flex-$\pi$, or, for TesserAct, as text, without alignment to pixels or scene points. EnerVerse-AC~\cite{jiang2025evac} and MTV-World~\cite{su2025mtvworld} project the end-effector pose or trajectory into the camera view, and PointWorld~\cite{huang2026pointworld} represents the gripper motion as 3D point flows and regresses the displacement of the observed scene points in one-second chunks. All three ground the action spatially but leave its effect on the scene to the network. ClothCast also applies an analytic prior to the cloth points themselves, carrying every point with the grippers that hold the cloth, and learns only the remaining deformation, in the spirit of residual physics~\cite{zeng2020tossingbot}.

\paragraph{Cloth dynamics models.}
Models built for cloth act on an explicit state: MeshGraphNets~\cite{pfaff2021meshgraphnets} prescribes the motion of mesh nodes, VCD~\cite{lin2021vcd} and PGND~\cite{zhang2025pgnd} constrain the grasped particles to the gripper, PhysTwin~\cite{jiang2025phystwin} drives a fitted spring-mass simulator, and UniClothDiff~\cite{tian2025uniclothdiff} conditions a diffusion dynamics model on the actions. They need a supplied mesh (MeshGraphNets), a category-specific reconstruction (GarmentNets~\cite{chi2021garmentnets}, MEDOR~\cite{huang2022medor}), a garment template with a history of estimated states (UniClothDiff, concurrent work~\cite{sim2026incontext}) or a fit to an interaction video of the object (PhysTwin); VCD and PGND need only what the cameras see, but add no unobserved material, which ClothCast hypothesizes from a single observation with one model for all categories.

\paragraph{Cloth datasets.}
Simulators such as SoftGym~\cite{lin2020softgym} and GarmentLab~\cite{lu2024garmentlab} target policy learning; closest to ours, VR-Folding~\cite{xue2023garmenttracking} pairs multi-view RGB-D with a complete garment mesh at every frame of human VR folding. ClothCastBench adds dual-arm gripper trajectories and per-pixel labels for up to six cloth layers, with their correspondence over time.

\section{Method}
\label{sec:method}

\subsection{Setting and Overview}
\label{sec:overview}

\paragraph{Setting.}
At inference, ClothCast predicts the RGB-D observations of a whole episode from only the state observable at frame~0 (time $t_0$) and the plan: one $256\times256$ RGB-D observation with the camera calibration, the robot's own model (kinematics and meshes), and the planned gripper trajectories of both arms, a pose and a binary closed flag per frame at 10\,Hz. The camera pose is known when the camera is mounted on a tracked rig or on the robot, and can otherwise be recovered from the $t_0$ frame, as we show in simulation (\cref{sec:exp_ablation}). The calibration and the robot model serve only geometry, including a robot mask for every frame rendered from the meshes posed by inverse kinematics on the planned trajectories; at $t_0$, the pixels where the observed depth lies in front of the rendered arm are removed from it. Nothing else about the cloth is given: no mesh, template, category or material.

\paragraph{Overview.}
The cloth state is a set of 3D points: a rigid prior moves them with the grippers that hold the cloth (\cref{eq:prior}), and Stage~2 learns the residual. Since Stage~2 never creates points, Stage~1 lifts the observation to a layered point set that already contains the occluded cloth; Stage~3 renders the moved points to depth and RGB. Only the three diffusion models are learned, not the prior, the settling after release or the depth rendering.

\paragraph{Shared architecture.}
The three stages share one architecture and are trained separately on ClothCastBench (\cref{sec:benchmark}). Every map a stage reads or writes, except Stage~2's gripper heat maps, which enter at latent resolution, is a $256\times256$ image in the camera view, encoded by the frozen Stable Video Diffusion (SVD)~\cite{blattmann2023stable} VAE into a $4\times32\times32$ latent tile; as in DepthWorld~\cite{depthworld}, tiles are placed side by side to form each stage's canvas. The denoiser is the SVD UNet as extended by Ctrl-World~\cite{guo2026ctrl}, with history-frame and frame-level conditioning and an action encoder. Stages~1 and~3 feed the action encoder a constant and start from the SVD weights; Stage~2, the only stage in which the action input varies, starts from DepthWorld's released UNet.

\subsection{Stage 1: Layered Geometry at $t_0$}
\label{sec:stage1}

Stage~1 estimates the cloth geometry at $t_0$ from the single RGB-D observation as a layered depth image~\cite{shade1998ldi}, a representation previously predicted from a single view for view synthesis~\cite{tulsiani2018layer,shih2020photo}: for each pixel, the depths of up to four cloth surfaces that its camera ray crosses. The conditioning canvas holds only the RGB image, the scene depth and a second copy of the RGB, matching the three generated tiles, and the history frame holds the same observation; no mask or other label enters the generator. It generates three tiles: the depth of the surface layer, the cloth surface nearest to the camera; a gap tile whose three channels give the depth offsets of up to three hidden layers, each beneath the one above it; and a visible-cloth mask. The surface layer is predicted, not read from the sensor, because it includes cloth behind the robot and pixels without valid sensor depth; together, the four layers can represent 87\% of the cloth-layer pixels labeled at $t_0$. The targets are the benchmark's layered depth labels (\cref{sec:benchmark}); at inference we keep the first generated frame of a single sample.

The predicted tiles are then combined with the observation. On pixels that the predicted mask marks as visible cloth, outside the robot mask and with valid sensor depth, the surface layer takes the observed depth; a scalar factor $k$, fitted by least squares between predicted and observed depth on these pixels, rescales the predicted surface everywhere else, and each hidden layer is placed beneath this surface by its $k$-scaled offsets. Unprojecting the resulting depth stack gives the layered point set $\hat{P}_0$, each point labeled with its layer, which is the only cloth geometry the rest of the pipeline receives.

\subsection{Stage 2: Action-Conditioned Point Motion}
\label{sec:stage2}

\begin{figure}[t]
  \centering
  \includegraphics[width=0.7\linewidth]{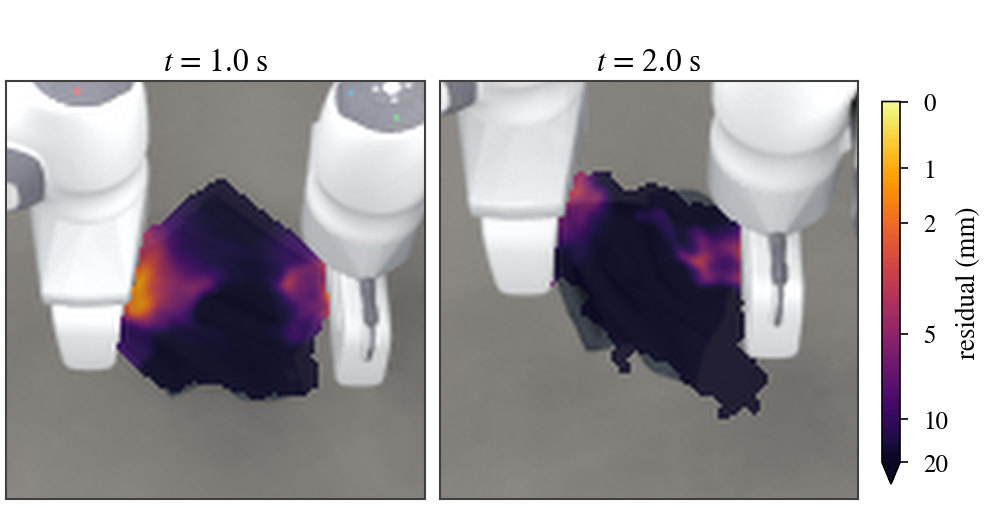}
  \caption{\textbf{Stage-2 residual over the rigid prior.} Distance between each pixel's front-most predicted point and its rigid-prior position (\cref{eq:prior}) on the episode of \cref{fig:pipeline}, at two frames of the first window; a departure from the prior, not an error (bright: small; nonlinear scale saturating at 20\,mm; robot pixels uncolored). The median is about 5\,mm within 3\,cm of a holding gripper, and 18\,mm at 1\,s and 73\,mm at 2\,s for points 10 to 15\,cm away, where the hanging cloth departs from the grippers' rigid motion; the pull toward the prior (\cref{sec:stage2}) affects under 3\% of the colored pixels.}
  \label{fig:residual}
\end{figure}

Stage~2 moves the point set through the episode in windows of 21 frames (2\,s): the first frame of each window is its history frame, the model predicts the following 20, and rollouts chain windows from the points predicted at the end of the previous one.

\paragraph{Rigid prior.}
Let $T^a_t$ and $R^a_t$ be the position and orientation of arm $a$'s gripper at frame $t$ of the window, with $t=0$ its first frame, $c^a_t$ its closed flag, and $p_i$ the position of point $i$ at $t=0$. Each arm moves a point rigidly by $\Phi^a_t$, and the prior blends the two motions:
\begin{equation}
\begin{aligned}
\bar{p}_{i,t} &= \sum_{a} w^a_{i,t}\, \Phi^a_t(p_i), \\
\Phi^a_t(p) &= R^a_t (R^a_0)^{\top} (p - T^a_0) + T^a_t, \\
w^a_{i,t} &= \frac{g^a_t\, e^{-\lVert p_i - T^a_0 \rVert/\lambda}}{\sum_b g^b_t\, e^{-\lVert p_i - T^b_0 \rVert/\lambda}},
\end{aligned}
\label{eq:prior}
\end{equation}
where $g^a_t = c^a_0\, c^a_t$ marks an arm that holds the cloth at both the window start and frame $t$, and $\lambda = 0.05$\,m, which matters only while both arms hold. When one arm holds, its weight is one and the whole point set moves as a rigid body with that gripper; when neither arm holds, both weights are $\tfrac12$, so the prior averages the two grippers' rigid motions, and a grasp that closes inside a window enters the prior from the next window. The prior never keeps a point static: cloth resting on the table while the grippers move is carried by the prior, and the residual must return it. \cref{sec:exp_ablation} evaluates the prior on its own.

\paragraph{Residual flow.}
Motion is written as flow tiles, one per layer and frame: the pixel at which layer $\ell$ shows point $i$ at $t=0$ stores that point's cumulative displacement since $t=0$, in camera axes. Per frame, the conditioning canvas holds the window start's RGB, surface-depth and hidden-gap tiles and four prior tiles, $\bar{p}_{i,t} - p_i$ encoded over $\pm0.3$\,m; the model generates the same seven tiles, the first three supervised with that frame's ground-truth tiles, and only its four residual tiles are used. Each stores $r_{i,t} = p_{i,t} - \bar{p}_{i,t}$, with $p_{i,t}$ the point's ground-truth position from the benchmark's correspondence labels (\cref{sec:benchmark}), divided per frame by $3\sigma_t$, $\sigma_t=\max(0.2\,\mathrm{m}, \max_a \lVert T^a_t - T^a_0 \rVert)$ over the holding arms (0.2\,m if none), and clipped to the tile range, so that small and large manipulations occupy the same signal range. At inference, the conditioning RGB tile is always the observed image at $t_0$ (in training, the window's first image), while the surface and gap tiles are re-rendered from the current point set at the start of each window: each point takes the pixel it projects to and a layer by its depth order there, and points beyond the fourth layer or outside the image move with the prior alone.

The plan also enters in two further forms: per frame and per arm, the gripper position, the grasp anchor (the gripper position when the current grasp closed, zero while open) and the closed flag are embedded into cross-attention tokens by an action encoder of the Ctrl-World design~\cite{guo2026ctrl}; and to give the network an explicit spatial cue for where the cloth is held, the gripper positions and grasp anchors also enter as $32\times32$ heat-map channels per frame (supplement).

\paragraph{Decoding.}
At decode time the residual is added back to the prior and applied to the points. Because cloth held by a closed gripper moves with it, the surface-layer flow is also pulled toward the prior in a small neighborhood of each holding gripper: at every frame in which the arm holds, a pixel at distance $d$ (in pixels) from the gripper's position in the image at the window's first frame takes $(1-w)$ times the predicted flow plus $w$ times the prior, with the Gaussian weight $w=\exp(-d^2/32)$, which is 1 at the gripper and below 0.15 beyond 8 pixels. \cref{fig:residual} shows the predicted residual on a test episode.

\paragraph{Decoded-space supervision.}
The latent diffusion loss weights every pixel equally, so the few pixels of cloth near a gripper contribute little to it, although an error there separates the held cloth from its gripper. We therefore add a loss on decoded tiles: at each training step, one predicted frame and one flow tile (the surface layer with probability $\tfrac12$, otherwise a random hidden layer) are decoded with the frozen VAE from both the prediction and the ground-truth latent, so that the VAE's own errors do not count against the model. With $\delta$ the per-pixel, channel-averaged absolute difference between the two decoded tiles, $\mathcal{L}_{\text{flow}}$ averages $\delta$ over the decoded layer's occupancy at the window's first frame, and $\mathcal{L}_{\text{anchor}}$ averages it on the surface tile under that frame's gripper-position heat map (maximum over both grippers, normalized to one), only on frames where an arm is closed. Each term is rescaled to the running magnitude of the diffusion loss and capped (supplement).

\paragraph{Post-release settling.}
After the last frame in which any arm holds the cloth, the cloth falls with no planned motion to follow and the learned motion is less accurate (\cref{sec:exp_ablation}), so we replace it by settling the point set onto the table, whose height is read from the observed $t_0$ depth around the predicted cloth mask, outside the robot mask. From the release frame on, every point keeps its horizontal position, and the points of each 8-mm cell of a horizontal grid are stacked, in their height order at release, over a height proportional to their number and to $h_0$, the cloth's median height above the table at $t_0$, so denser cells pile higher, as folded cloth does; each point's height relaxes exponentially toward its target, with a time constant of 0.15\,s.

\subsection{Stage 3: RGB-D Completion}
\label{sec:completion}

\paragraph{Depth rendering.}
Projected directly into a depth image, the predicted points leave holes where the cloth has stretched, and isolated depth outliers. A gap fill before projection inserts points, on pixels left empty, along stretched links between neighboring points of the frame-0 point set (supplement); after projection, which keeps the nearest depth per pixel, a depth repair keeps the main connected cloth region, replaces depth spikes and fills the remaining holes inside it. The depth repair and the settling of \cref{sec:stage2} form the post-processing that \cref{sec:exp_ablation} ablates. The result is the predicted depth sequence, and the same surface, unprojected, is the cloth geometry we evaluate.

\paragraph{Depth-to-RGB synthesis.}
The completion network synthesizes the RGB frames and sees the plan only through its conditioning. Its conditioning canvas is a single tile per frame: the predicted depth of the cloth surface alone, log-encoded over 0.2--1.5\,m, with the robot mask of \cref{sec:overview} painted in a fixed color. Rollouts use windows of one context frame and four predicted frames, each conditioned on its own canvas; the first window's context is the observed RGB frame at $t_0$, and every later window takes its context from the model's own last predicted frame. During training, the canvases are built from the ground-truth cloth surface depth with the dataset's robot masks painted in.

\section{ClothCastBench}
\label{sec:benchmark}

ClothCastBench provides the geometry of occluded cloth and the motion of every cloth point for 10{,}000 simulated episodes in which two Franka Panda arms manipulate one cloth item, 2{,}500 for each of four classes: square sheets and procedurally generated t-shirts, long-sleeve shirts and pants. The cloth is simulated as a mass-spring system with self-collision and table friction, using GPU kernels adapted from PhysTwin~\cite{jiang2025phystwin}, and rendered with Blender's EEVEE engine; each item has one of ten fabric classes, which set its stiffness, mass, friction and texture. Every episode starts from a crumpled state and runs one of three scripted, open-loop dual-arm programs on two exposed grasp points: an inward page fold, with the arms synchronized or staggered (4{,}000 and 2{,}000 episodes), or a lift--swivel--carry--drop (4{,}000); each closed gripper attaches the cloth vertices within 3\,cm of it by virtual springs. A fixed pseudo-random split by episode holds out 516 test episodes and keeps 9{,}484 for training. Construction details are in the supplement.

Each episode provides $256\times256$ RGB-D frames at 10\,Hz from a calibrated camera, fixed within an episode and placed differently in each, together with visible-cloth and robot masks. The simulator adds, at every frame, the ground-truth cloth mesh and a layered depth decomposition of the camera view: at each pixel, the depth of the nearest cloth surface, including cloth hidden behind the robot, and of up to five cloth layers beneath it, each carrying the mesh face index and barycentric coordinates of the surface point it images. A pixel on any layer of any frame therefore identifies a fixed point on the cloth, whose ground-truth position is known at every later frame. Each episode also records, per frame and per arm, the gripper pose and a binary closed flag, and a 14-D action of per-arm increments in gripper translation, rotation and closure. Episodes are replayed kinematically, so the planned robot state is also the achieved one, and the action stream holds only gripper states; the cloth enters it only through the grasp points, which the scripted programs choose on exposed cloth. The arms are posed by inverse kinematics on the gripper trajectories, and arm--cloth contact is not simulated: the robot acts on the cloth only through the grasp springs, so another arm reaching the same gripper poses can be rendered in its place. The cloth mesh, the layered depth, the visible-cloth mask, and all frames after $t_0$ are used only for training and evaluation.

\section{Experiments}
\label{sec:experiments}

\subsection{Setup}
\label{sec:exp_setup}

\paragraph{Test set and protocol.}
We evaluate on the 516 held-out test episodes, which span the four classes and last 44--110 frames. Cosmos-Predict2.5 produced no output for 24 of them, so all comparisons use the remaining 492 episodes. The split is by episode, and no test episode is used to train any of the compared models. Each method starts from the observation at $t_0$ (RGB-D for ClothCast, Flex-$\pi$ and DepthWorld, RGB otherwise) and rolls out the whole episode autoregressively, conditioned on the planned gripper trajectory: the baselines receive it as the benchmark's 14-D actions, with the initial gripper state for Flex-$\pi$ and the gripper state at every frame for V-JEPA~2-AC, and ClothCast as gripper poses and grasp states. None of the baselines receives the camera pose or the robot model, and only Flex-$\pi$ receives the camera intrinsics, which all episodes share.

\paragraph{Inputs.}
The robot's kinematics give Stage~3 the arm silhouette along the planned trajectory (\cref{sec:overview}), whereas the baselines must generate the arm themselves. Stage~2 receives neither the robot mask nor the robot model at any frame, only the planned gripper trajectory, which the camera pose places in camera axes; the mask enters the geometry only at $t_0$, where it keeps arm pixels out of the observed cloth depth and the table-plane estimate. We therefore compare on the cloth. The geometry metric and cloth PSNR exclude the robot pixels, and cloth SSIM windows reach them only at the region boundary; the cloth LPIPS crop is on average 22\% robot, and pasting the true arm into ClothCast's frames lowers its cloth LPIPS by only about 0.005; we did not test how much of the baselines' cloth LPIPS comes from the arm they must generate. The full-frame metrics are not a like-for-like comparison, since the arm fills 19\% of the frame and ClothCast is given its silhouette; the geometric and RGB cloth metrics also favor ClothCast wherever a baseline draws the arm over cloth.

\paragraph{Metrics.}
For geometry, we measure the Chamfer distance on the visible cloth surface: in each predicted frame, the ground-truth visible-cloth pixels, excluding the robot, eroded twice with a $3\times3$ kernel and restricted to valid ground-truth depth, are back-projected from the prediction and from the ground truth, and the Chamfer distance is the mean of the two directed mean nearest-neighbor distances, in millimeters, without subsampling or alignment. Predicted cloth outside these pixels is not scored, so the metric does not penalize cloth predicted where the ground truth shows none. An evaluation pixel left without a predicted depth is not filled: its ground-truth point is matched to the nearest predicted point elsewhere on the surface, so every uncovered pixel adds its distance to the predicted surface. We report the coverage, the share of evaluation pixels with a predicted depth; the baselines predict every pixel. For RGB, we report PSNR, SSIM and LPIPS (VGG)~\cite{zhang2018lpips} on the full frame and on the same cloth region, where SSIM is averaged over the region and LPIPS is computed on a crop around it. Every metric is averaged over the frames of an episode, then over the episodes of a garment class, then over the four classes, excluding the observation at $t_0$.

\paragraph{Baselines.}
We compare against four action-conditioned world models, each scored on its own output without our depth repair or settling. Flex-$\pi$~\cite{flexpi} and DepthWorld~\cite{depthworld} predict RGB and per-pixel geometry, so they enter both comparisons; Cosmos-Predict2.5~\cite{cosmos} (Cosmos in \cref{tab:rgb}) and V-JEPA~2-AC~\cite{vjepa2ac} predict RGB only, V-JEPA~2-AC through an RGB decoder trained on the benchmark. Each baseline starts from its released checkpoint and is adapted to the 9,484 training episodes. Every 500 updates, a checkpoint is scored on a fixed probe of 20 training episodes, by cloth Chamfer distance for Flex-$\pi$ and DepthWorld and by cloth LPIPS for Cosmos and the V-JEPA~2-AC predictor; training was stopped manually once these scores had leveled off, and the V-JEPA~2-AC decoder was trained for the full 10 epochs. The supplement gives the probe protocol, every probe score and the training details. No test episode is used for training or checkpoint selection. MEDOR~\cite{huang2022medor} predicts pick-and-place outcomes rather than observations along a trajectory, and UniClothDiff~\cite{tian2025uniclothdiff} needs a garment template and a history of estimated states, so we do not compare against either. PointWorld~\cite{huang2026pointworld} shares our inputs but forecasts only the observed points, in 1-second chunks and without images; we do not retrain it on ClothCastBench, and dropping our hidden-layer points (\cref{sec:exp_occluded}) only approximates its restriction to observed points.

\paragraph{Implementation details.}
Each stage trains its 1.53B parameters (VAE frozen) with AdamW at a constant learning rate of $10^{-5}$ and an effective batch of 4, and uses EMA weights (decay 0.999) and 25 denoising steps (guidance schedule in the supplement). Stage~1 trains on 1,498 episodes of an earlier 1,744-episode subset of the benchmark, Stage~2 on all 9,484 training episodes and Stage~3 on 8,998 of them (clip counts in the supplement); for Stage~2 we use the checkpoint after 16,000 updates (1.5 epochs), at the end of a plateau that began at 14,000.

\subsection{Comparison with world models}
\label{sec:exp_sim}

\begin{table}[t]
\centering
\small
\setlength{\tabcolsep}{4.5pt}
\begin{tabular}{@{}lcccccc@{}}
\toprule
& \multicolumn{6}{c}{Chamfer distance (mm) $\downarrow$} \\
\cmidrule(l){2-7}
Method & 1\,s & 2\,s & 3\,s & 4\,s & Last & All \\
\midrule
DepthWorld & 17.9 & 48.1 & 77.8 & 70.5 & 88.7 & 61.1 \\
Flex-$\pi$ & 21.1 & 39.6 & 49.3 & 35.0 & 37.4 & 35.6 \\
ClothCast & \textbf{8.2} & \textbf{14.9} & \textbf{17.3} & \textbf{16.7} & \textbf{16.2} & \textbf{14.0} \\
\midrule
\textit{ClothCast coverage} & \textit{0.98} & \textit{0.91} & \textit{0.86} & \textit{0.83} & \textit{0.81} & \textit{0.89} \\
\bottomrule
\end{tabular}
\caption{\textbf{Visible-surface geometry} on the 492 test episodes, which all last beyond 4\,s; Last: each episode's final frame, after release, where settling (\cref{sec:stage2}) places ClothCast's points. The baselines predict every pixel (coverage 1.00).}
\label{tab:geometry}
\end{table}

\begin{table}[t]
\centering
\small
\setlength{\tabcolsep}{3.2pt}
\begin{tabular}{@{}lcccccc@{}}
\toprule
& \multicolumn{3}{c}{Cloth} & \multicolumn{3}{c}{Full frame} \\
\cmidrule(lr){2-4}\cmidrule(l){5-7}
Method & PSNR & SSIM & LPIPS & PSNR & SSIM & LPIPS \\
\midrule
DepthWorld & 14.99 & 0.340 & 0.124 & 18.06 & 0.758 & 0.064 \\
Flex-$\pi$ & 15.16 & 0.347 & 0.132 & 16.84 & 0.737 & 0.070 \\
V-JEPA~2-AC & 17.11 & 0.393 & 0.130 & 19.20 & 0.786 & 0.060 \\
Cosmos & 17.21 & 0.381 & 0.114 & 20.01 & 0.791 & 0.052 \\
ClothCast & \textbf{20.93} & \textbf{0.456} & \textbf{0.076} & \textbf{28.59} & \textbf{0.934} & \textbf{0.021} \\
\bottomrule
\end{tabular}
\caption{\textbf{RGB} on the 492 test episodes, averaged over all predicted frames; PSNR and SSIM: higher is better, LPIPS: lower is better. The full frame includes the robot, whose silhouette ClothCast receives (\cref{sec:exp_setup}) and the baselines must generate.}
\label{tab:rgb}
\end{table}

\paragraph{Geometry.}
ClothCast's visible-surface Chamfer distance averages 14.0\,mm over all predicted frames, 2.5$\times$ lower than Flex-$\pi$, the strongest baseline in geometry (35.6\,mm), and 4.4$\times$ lower than DepthWorld (61.1\,mm), and it is lower at every horizon (\cref{tab:geometry}). Filling every uncovered pixel with the planned arm's depth where the arm covers it and the $t_0$ background elsewhere gives 14.0\,mm over all frames, and filling it with the nearest predicted depth gives 14.5\,mm. The ground-truth-to-prediction term, which includes the uncovered pixels, is 13.0\,mm for ClothCast, against 22.0\,mm for Flex-$\pi$ and 39.0\,mm for DepthWorld.

\paragraph{RGB.}
ClothCast reaches 20.93\,dB cloth PSNR, 3.7\,dB above Cosmos-Predict2.5, the strongest baseline in cloth PSNR, with the highest cloth SSIM and the lowest cloth LPIPS (\cref{tab:rgb}). At the first predicted frame, Flex-$\pi$, Cosmos and DepthWorld have higher cloth PSNR (31.1, 31.0 and 29.1 vs.\ 28.2\,dB); from 1\,s on, ClothCast leads, by 4.9\,dB at the last frame (per-horizon values in the supplement). The full-frame margin is larger (28.59 vs.\ 20.01\,dB); we do not separate how much of it comes from the arm silhouette that ClothCast receives.

\subsection{Occluded geometry}
\label{sec:exp_occluded}
\begin{table}[t]
\centering
\footnotesize
\setlength{\tabcolsep}{5pt}
\begin{tabular}{@{}lccc@{}}
\toprule
Predicted points & Full & Occluded & Hidden pts \\
\midrule
Surface layer only & 16.4 & 14.2 & -- \\
All four layers & \textbf{13.4} & \textbf{9.4} & 15.2 \\
\bottomrule
\end{tabular}
\caption{\textbf{Occluded geometry} (mm, all frames, 492 test episodes): Chamfer distance to the ground-truth cloth on all layers (Full); distance from occluded ground-truth cloth to the nearest predicted point (Occluded) and from hidden-layer points to the nearest ground-truth cloth (Hidden pts).}
\label{tab:occluded}
\end{table}
The benchmark's layered labels let us score the occluded layers directly. On the rollouts of \cref{sec:exp_sim}, we compare the moved points with the same points restricted to the Stage-1 surface layer, both against the ground-truth cloth on all layers (\cref{tab:occluded}); we do not train a model without the hidden layers. With them, the full-geometry Chamfer distance falls from 16.4 to 13.4\,mm (paired difference 3.0\,mm, 95\% interval 2.9 to 3.1), and occluded cloth lies 9.4 instead of 14.2\,mm from the nearest predicted point (9.1 instead of 14.3\,mm for cloth hidden from $t_0$ on), closer in every episode. Over all frames, 20\% of the hidden-layer points come to the front, against 17\% of the cloth hidden at $t_0$ that becomes visible in the ground truth.

\subsection{Ablations}
\label{sec:exp_ablation}
We ablate the learned residual and the post-processing, and test whether ClothCast's advantage depends on a measured camera pose, on 50 randomly sampled, class-balanced test episodes, rolled out on a different GPU and with different random seeds than \cref{sec:exp_sim}, under which ClothCast scores 14.5\,mm (14.0\,mm in the main run on these episodes, a run-to-run variation from sampling and hardware); variants that leave Stage~2's inputs unchanged reuse its samples, and 95\% intervals are bootstrapped over paired episode differences.

\begin{table}[t]
\centering
\footnotesize
\setlength{\tabcolsep}{2pt}
\begin{tabular}{@{}ccccccc@{}}
\toprule
& \multicolumn{2}{c}{w/o post-processing} & Raw & \multicolumn{2}{c}{Prior only} & Camera \\
\cmidrule(lr){2-3}\cmidrule(lr){5-6}\cmidrule(l){7-7}
ClothCast & settle & repair & points & post. & raw & from $t_0$ \\
\midrule
14.5 & 15.5 & 14.7 & 15.8 & 21.2 & 24.8 & 14.3 \\
\textit{0.90} & \textit{0.90} & \textit{0.87} & \textit{0.76} & \textit{0.81} & \textit{0.72} & \textit{0.90} \\
\bottomrule
\end{tabular}
\caption{\textbf{Ablations}: mean Chamfer distance (mm) over all predicted frames on the 50 ablation episodes. Second row: coverage. \emph{w/o}: the named step removed. \emph{Raw points}: no gap fill or post-processing. \emph{Prior only}: learned residual set to zero (prior of Eq.~\eqref{eq:prior} chained over the same windows), with (post.) or without (raw) gap fill and post-processing. \emph{Camera from $t_0$}: pose recovered from the $t_0$ frame. Per-horizon values are in the supplement.}
\label{tab:ablation}
\end{table}

\paragraph{Learned motion.}
The rigid prior alone reaches 21.2\,mm (\cref{tab:ablation}), already below both geometry baselines on these episodes (34.2\,mm for Flex-$\pi$ and 67.2\,mm for DepthWorld); the learned residual lowers this to 14.5\,mm and raises the coverage from 0.81 to 0.90. Without post-processing and relative to Flex-$\pi$ (34.2\,mm), moving the Stage-1 points with the prior (24.8\,mm) and adding the learned residual (15.8\,mm) each account for about half of the reduction; the prior differs from the baselines both in its representation and in its use of the camera pose, which can be recovered from the $t_0$ frame with the robot model, as shown below. The gain is largest while the cloth is manipulated (31.1 to 15.7\,mm at 2\,s, 28.2 to 19.1\,mm at 3\,s) and small at the last frame, after settling has replaced the predicted motion in both (15.9 vs.\ 15.8\,mm).

\paragraph{Post-processing.}
Settling acts after release: removing it leaves the error unchanged up to 3\,s but raises the last-frame error from 15.8 to 20.8\,mm. The depth repair mainly completes the rendered surface (coverage 0.87 without it). The rigid prior depends more on post-processing: it moves cloth only with the grippers, which stay still after release in every test episode, so released cloth stays in the air instead of falling, and its raw points end at 36.5\,mm instead of 15.9\,mm.

\paragraph{Camera pose from the first frame.}
Where the camera pose is not measured, ClothCast recovers it from the $t_0$ frame: the table plane in the depth fixes the camera's tilt and height, and a yaw search followed by ICP registers the arms' model, posed at its planned $t_0$ configuration, to the points above the table. On 514 of the 516 test episodes, the recovered pose is within 0.17$^\circ$ and 1.3\,mm of the calibration (median 0.04$^\circ$ and 0.5\,mm); the two failures are a wrong table plane and a 180$^\circ$ ambiguity from symmetric arm poses. On the ablation episodes, the recovered pose gives a Chamfer distance of 14.34\,mm (\cref{tab:ablation}), against 14.47\,mm with the calibration (paired difference $-0.13$\,mm, 95\% interval $-0.52$ to 0.15\,mm). In simulation, with noise-free depth, ClothCast's margin over the world models thus does not require a measured camera pose, which Stage~2 uses to place the gripper trajectory in its heat maps and rigid prior; the recovery itself needs the robot model, which the world models do not receive, and is untested on real data.

\begin{figure}[t]
    \centering
    \includegraphics[width=0.8\linewidth]{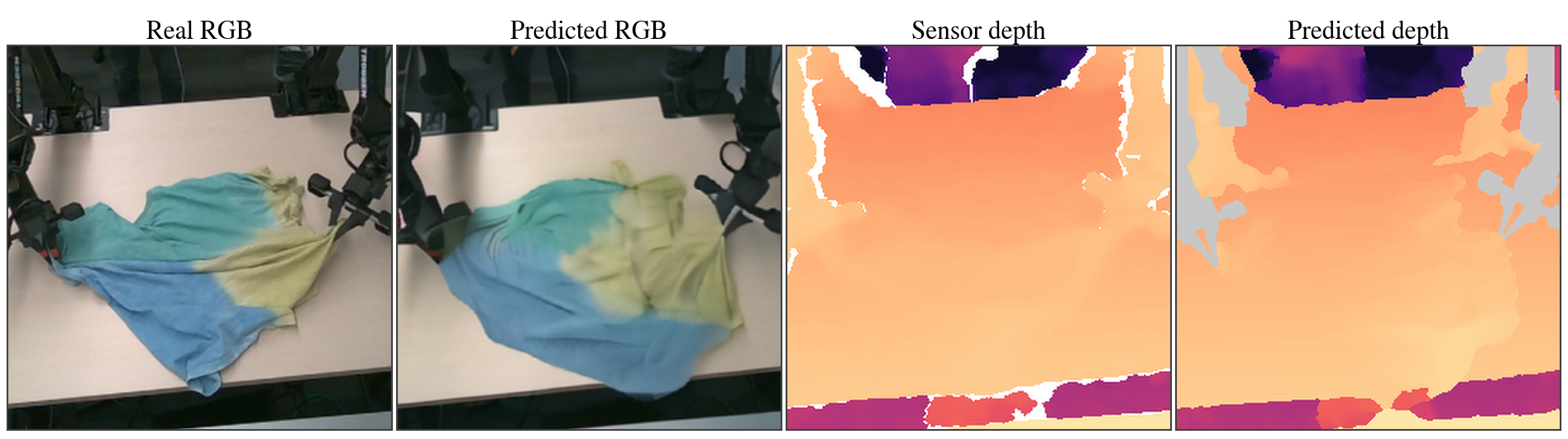}
    \caption{\textbf{Real test episode} (t-shirt, two-arm drag), 2.0\,s after $t_0$, chosen among the frames with the highest cloth PSNR at least 2\,s after $t_0$: observed and predicted RGB, sensor and predicted depth (gray: the arm, rendered from the measured joints).}
    \label{fig:real}
\end{figure}

\subsection{Real-robot evaluation}
\label{sec:exp_real}
Our real data has 418 episodes in which two WidowX~AI arms carry, drag, lift, pull apart or swing seven items (two gingham cloths, a towel, a polo shirt, a t-shirt and two pillow covers), recorded as $256\times256$ RGB-D at 10\,Hz by a RealSense D455 with the camera calibration, the measured gripper trajectory and approximate masks of the visible cloth (grown from color) and of the robot (rendered from the measured joints), but no cloth mesh. We hold out 84 random episodes for testing (\cref{fig:real}); the same items appear in the 334 training episodes.

Stage~2 is used as trained in simulation, and Stage~1 is fine-tuned only on simulated training episodes re-rendered with WidowX arms, so neither geometry stage is trained on a real image. Stage~3 is fine-tuned for three epochs on the real training episodes, conditioned on the sensor depth of the visible cloth and the robot mask. Two inference settings differ from simulation: the forecasts are conditioned on the executed trajectory, measured rather than commanded, and the reference height $h_0$ of settling is fixed at 20\,mm, chosen on a training episode on which the value computed from the Stage-1 points (44\,mm) stacked the released cloth up to 0.36\,m high. As in simulation, the future robot masks given to Stage~3 are rendered from the joints alone, without the sensor depth.

Against the sensor depth, with the Chamfer distance of \cref{sec:exp_setup} averaged per item and then over the seven items, the predicted visible surface of the 84 test episodes has an error of 5.6\,mm at 0.1\,s, 32.6\,mm at 1\,s, 34.7\,mm at the last frame and 35.3\,mm over all frames, at a coverage of 0.66; the cloth PSNR is 14.6\,dB and the cloth LPIPS 0.137. The rigid prior alone (\cref{sec:exp_ablation}), with the same Stage-1 points and post-processing, reaches 5.0, 54.5, 30.8 and 44.9\,mm at the same horizons, at a coverage of 0.67. Over all frames, the learned residual thus lowers the error by 9.6\,mm; it helps while the cloth is moved and is worse than the prior at the first and the last frame. The error is lower on the items closest to the benchmark's classes, flat cloths and shirts: 30.2\,mm over all frames on the 64 episodes of the gingham cloths, the towel and the shirts, against 48.1\,mm on the two pillow covers, which are sewn two-layer sleeves and whose test episodes are mostly carries. Per-horizon and per-item results are in the supplement.

\section{Limitations}
\label{sec:conclusion}


ClothCastBench is a preliminary benchmark, not a deployment-ready one. It covers three scripted dual-arm programs, in which the robot acts on the cloth only through grasp springs, and needs more cloth types, motions, fabrics and real data. On real data, the error is highest on the two pillow covers, whose test episodes are mostly carries, and carries have the highest error for every item; the real RGB forecasts use a completion stage fine-tuned on footage of the same items. The four modeled layers leave 13\% of the labeled layer pixels at $t_0$ unrepresented, and the hidden-layer points come to the front in about the right amount but not reliably where hidden cloth is revealed: at any given horizon, they show at revealed and at previously visible pixels at similar rates. Settling, gap fill and depth repair are hand-designed (\cref{sec:exp_ablation}), and points drift along the surface over long horizons. Finally, each forecast is a single sample that takes minutes per episode, so planning with ClothCast, which needs several samples per candidate action sequence, is left to future work.
\section*{LLM Usage Statement}
We used Claude (Anthropic) through the Claude Code agent, and ChatGPT (OpenAI) in this work.
\emph{Code and experiments:} Claude Code agent wrote and modified much of the code for data conversion, training, inference, post-processing and evaluation under the authors' extensive supervision and direction.
\emph{Analysis:} Claude Code agent computed and aggregated the reported metrics, re-implemented and cross-checked the evaluation metric, and audited the data split and the inputs each method receives.
\emph{Writing:} Claude and ChatGPT helped refine the text in every section, including the related work and the description of the benchmark generator, produced the \LaTeX{} tables, and served as mock reviewer for iterations on the draft. Claude Code agent helped write the code for the figures in the paper.
\emph{Ideation:} Claude and ChatGPT helped with part of the related-work search, and provided preliminary feedback for experiments.
All LLM output was reviewed by the authors, who take full responsibility for the code, the results and the text.

\maketitlesupplementary

\setcounter{section}{0}
\setcounter{table}{0}
\setcounter{figure}{0}
\setcounter{equation}{0}
\renewcommand{\thesection}{\Alph{section}}
\renewcommand{\thetable}{S\arabic{table}}
\renewcommand{\thefigure}{S\arabic{figure}}
\renewcommand{\theequation}{S\arabic{equation}}

\noindent Section, table and equation numbers without a letter refer to the main paper. \Cref{sec:supp_benchmark} details the construction of ClothCastBench (Sec.~4), \cref{sec:supp_method} the implementation of the three stages, the settling and the depth rendering (Sec.~3), \cref{sec:supp_metrics} the metrics (Sec.~5.1), \cref{sec:supp_baselines} the inputs and training of the baselines with every probe score, \cref{sec:supp_horizon} the per-horizon results of Secs.~5.2 and~5.4, and \cref{sec:supp_real} the real-robot evaluation (Sec.~5.5).

\section{ClothCastBench construction}
\label{sec:supp_benchmark}

This section completes Sec.~4; every value is read from the generator's code and the released episode files.

\paragraph{Items.}
Sheets are $31\times31$-vertex grids with 1\,cm spacing. T-shirts (short sleeves), shirts (long sleeves, a collar band and an overlapping front placket) and pants (two legs) consist of masked quad-grid panels with 1\,cm spacing, whose front panels are stitched to the back panel along the outline except at the openings. Their dimensions are drawn uniformly from ranges of real garments and then halved; before halving, t-shirts have a body 34--44\,cm wide and 45--58\,cm long with 10--16\,cm sleeves, shirts a body 38--46\,cm wide and 58--70\,cm long with 48--56\,cm sleeves, and pants a 32--40\,cm waist and a length of 60--74\,cm. Every panel has structural, shear and two-hop bending springs.

\paragraph{Simulation.}
The mass-spring kernels, adapted from PhysTwin~\cite{jiang2025phystwin}, integrate 167 substeps per 30\,Hz frame ($\Delta t\approx0.2$\,ms) with dashpot and drag damping. Cloth--cloth and cloth--table contacts are inelastic, with the fabric class's Coulomb friction; the cloth--cloth contact distance stands in for ply thickness, and self-collision ignores vertex pairs that are close in the rest shape. \Cref{tab:supp_fabric} lists the ten fabric classes. Stiffnesses are log stiffness-to-mass ratios in the simulator's units, the log dashpot damping is drawn from 3.5--4.8 for every class, and the mass is scaled by the item's areal density relative to 150\,g/m$^2$, clamped to $[0.4, 3.5]$. The class names indicate which fabrics the ranges imitate; the values are not calibrated material constants, and masses are not in SI units.

\paragraph{Grasps.}
The cloth vertices ``within 3\,cm'' of a closed gripper (Sec.~4) are those inside a vertical cylinder of 3\,cm radius around the grasp point, reaching 2.5\,cm below it. Control springs attach the gripper to the connected patch that contains the grasp point and to the patches of lower layers inside the cylinder; a lower patch is released once its pull exceeds a friction limit. The two cylinders may not share vertices, and an opened gripper does not grasp again.

\paragraph{Initial states.}
The initial states are the starting states of an earlier dataset generated with the same simulator and garment generator, in which each flat item was folded by page-turn arcs about a crease and crumpled by lifting it at one point and releasing it in the air; shirts begin with their sleeves folded across the body. They are used with zero velocity. In plastic items (47\% of the episodes), bending springs that were compressed below 75\% of their rest length during this preparation keep the compressed length, so the creases persist; shirts always keep the creases of their folded sleeves. A state qualifies if, relative to the flat item, its top-view footprint is at most 0.6 times as large and its overlap factor, the summed top-view area of its triangles divided by the footprint, at least 1.35 times as large; if the area-weighted dispersion of its normals (one minus the largest eigenvalue of their second-moment matrix) is at least 0.12; if at least 12\% of its area is tilted by more than 45$^\circ$; and if the spread between the 5th and 95th percentiles of its area-weighted height is at least 5\% of the square root of the flat footprint.

\paragraph{Grasp selection and programs.}
Grasp candidates are the vertices visible from the earlier dataset's camera in the initial state; on shirts, only the interior of the back torso. A pair must be at least 12\,cm apart and at least $\max(18\,\text{cm}, d)$ apart in the flat item, where $d$ is its current distance; pairs far apart horizontally are preferred. Each program starts 0.2--0.5\,s into the episode. Along a page-fold arc with stroke $\mathbf{s}$ and height $H$, the grasp point advances horizontally by $\tfrac12(1-\cos\theta)\,\mathbf{s}$ and rises by $H\sin\theta$, for $\theta\in[0,\pi]$; the lift--swivel--carry--drop lifts both grippers vertically by $H$, pauses for 0.1--0.3\,s and moves them horizontally by their strokes. The heights $(H_1,H_2)$ of the two grippers are (14, 12), (16, 14), (18, 16) or (20, 18)\,cm, in 2,539, 3,559, 1,473 and 2,429 episodes, and the strokes are 20 and 18\,cm long. In the synchronized programs, the strokes are the two-circle solutions that turn the vector between the grippers by 45$^\circ$ (7,097 of the 8,000 episodes), 60$^\circ$ (298), 75$^\circ$ (453) or 90$^\circ$ (152) and set its length to 0.65--1.15 times its initial value, and at least 14\,cm; the solution that moves the pair toward the item's center is tried first. In the staggered page fold, the first gripper moves 20\,cm toward the second, ending at least 14\,cm from it, and the second then moves 18\,cm in the same direction, starting 0.1\,s after the first stops. Every program keeps the grippers at least 10\,cm apart and no farther apart than the grasp points are in the flat item, stays below 0.25\,m/s and 0.70\,m/s$^2$, changes the vector between the grippers by at least 14\,cm, and, if synchronized, moves both grippers together during at least 90\% of its active time. The first jaw opens 0.1\,s after the motion ends and the second 0.1 or 0.2\,s later, and the episode ends 1.2\,s after the second release, 4.3--10.9\,s after it began.

\paragraph{Arms.}
The Panda bases stand 1.2\,m apart on opposite sides of the table, facing each other. Inverse kinematics poses each arm at its gripper's planned pose, and a plan is accepted only if every pose is reached within tolerance and within the joint limits, with no sampled collision between the arms or with the table. Collisions within an arm and between the arms and the cloth are not checked; the arms do not act on the cloth in the simulation.

\paragraph{Rendering.}
Blender~4.0.2 renders every third 30\,Hz simulation state, giving 10\,Hz, with EEVEE at 24 samples, ambient occlusion and soft shadows, under the same three disk area lights and dim world light in every episode. Each fabric class has one CC0 physically based texture set (color, roughness and normal maps) from the ambientCG library. The intrinsics are fixed (focal length 351.7\,px, 40$^\circ$ vertical field of view). The 18 candidate viewpoints combine the two sides of the grasp pair, three azimuth offsets along it and elevations of 55, 65 and 75$^\circ$, which are selected for 8,358, 1,337 and 305 episodes. Each candidate is scored by the projected cloth area it sees after ray tests against the arms and the cloth, summed at the start, at 40\% and at 85\% of the episode.

\paragraph{Labels and validation.}
Along each pixel's ray, the first six cloth surfaces are labeled, and deeper ones are dropped. Every released episode passes the generator's validation of its arrays, which checks, among others, the ordering and sentinels of the layer labels, that each face--barycentric label reprojects to its pixel (largest error $4.2\times10^{-7}$\,px) and to its depth ($6.0\times10^{-8}$\,m), that the robot mask excludes the cloth, and that the actions equal the differences of the recorded gripper states (largest error $6.0\times10^{-8}$). An episode is re-planned with a new attempt when planning fails, the simulator's stability guard trips, the cloth leaves the rendered table, or an export stage fails repeatedly; 1,129 of the released episodes come from a second or later attempt. No episode is selected for its task outcome.

\paragraph{Split.}
The test episodes are those whose name has an MD5 hash whose first eight hexadecimal digits, read as an integer, are divisible by 20: 516 episodes, namely 127 sheets, 134 t-shirts, 125 shirts and 130 pants, or 203 synchronized and 105 staggered page folds and 208 lift--swivel--carry--drops.

\begin{table*}[t]
\centering
\small
\setlength{\tabcolsep}{4pt}
\begin{tabular}{@{}llcccccccr@{}}
\toprule
Fabric class & Items & Areal density (g/m$^2$) & $\log k_\text{stretch}$ & $\log k_\text{shear}$ & $\log k_\text{bend}$ & Contact (mm) & Friction & Plastic & Episodes \\
\midrule
Cotton, plain & S, T, Sh & 100--160 & 9.9--10.6 & 8.6--9.6 & 5.3--6.0 & 8 & 0.45 & 0.68 & 1,330 \\
Striped shirting & S, T, Sh & 100--150 & 9.9--10.5 & 8.6--9.5 & 5.3--5.9 & 8 & 0.45 & 0.68 & 1,302 \\
Flannel, plaid & S, T, Sh & 150--220 & 9.9--10.5 & 8.4--9.4 & 5.5--6.1 & 10 & 0.50 & 0.61 & 1,340 \\
Knit jersey & S, T & 150--260 & 9.3--9.9 & 8.0--8.8 & 4.7--5.4 & 10 & 0.60 & 0.12 & 1,111 \\
Silk, satin, polyester & S, Sh & 40--90 & 9.6--10.2 & 8.2--9.0 & 4.5--5.1 & 8 & 0.25 & 0.11 & 634 \\
Cotton twill, denim & S, Sh, P & 300--450 & 10.5--11.1 & 9.2--10.3 & 6.3--7.0 & 12 & 0.55 & 0.49 & 1,410 \\
Velvet, corduroy & S, P & 250--350 & 10.2--10.8 & 8.8--9.6 & 5.9--6.5 & 12 & 0.55 & 0.40 & 893 \\
Wool felt, fleece & S, P & 250--400 & 10.0--10.6 & 8.8--9.6 & 6.3--6.9 & 15 & 0.70 & 0.26 & 841 \\
Canvas, burlap & S, P & 350--500 & 10.7--11.3 & 9.4--10.3 & 6.5--7.2 & 15 & 0.60 & 0.58 & 834 \\
Quilted, padded & S & 300--420 & 10.3--10.9 & 9.0--9.8 & 6.1--6.8 & 15 & 0.50 & 0.28 & 305 \\
\bottomrule
\end{tabular}
\caption{\textbf{Fabric classes} of ClothCastBench. Items: S sheet, T t-shirt, Sh shirt, P pants. Stiffnesses are log stiffness-to-mass ratios of the stretch, shear and bending springs in the simulator's units; contact is the cloth--cloth contact distance; plastic is the share of the class's episodes with plastic creases. Ranges are sampled uniformly per item.}
\label{tab:supp_fabric}
\end{table*}

\section{Implementation details}
\label{sec:supp_method}

\paragraph{Training and sampling.}
Each stage trains its 1.53B parameters with AdamW at a constant learning rate of $10^{-5}$, an effective batch of 4 and fp16 mixed precision, keeps EMA weights (decay 0.999) for testing, and samples with 25 denoising steps and a guidance scale rising from 1.0 to 1.5 across the frames of a window. \Cref{tab:supp_stages} lists the training data and updates of each stage and of the two real-data fine-tunes of \cref{sec:supp_real}.

\begin{table}[t]
\centering
\footnotesize
\setlength{\tabcolsep}{3.5pt}
\begin{tabular}{@{}lllrr@{}}
\toprule
Stage & Initialized from & Clips & Updates & Epochs \\
\midrule
1 & SVD & 17,917 & 14,000 & 3.1 \\
2 & DepthWorld UNet & 43,684 & 16,000 & 1.5 \\
3 & SVD & 107,243 & 26,000 & 0.97 \\
\midrule
1, WidowX & Stage 1 & 15,778 & 4,000 & 1.0 \\
3, real & Stage 3 & 4,071 & 3,060 & 3.0 \\
\bottomrule
\end{tabular}
\caption{\textbf{Training of the three stages} and of the real-data fine-tunes. Clips have five frames (Stages~1 and~3) or 21 (Stage~2) and come from 1,498 training episodes of an earlier 1,744-episode subset of the benchmark (Stage~1), all 9,484 training episodes (Stage~2), 8,998 of them (Stage~3), 1,326 simulated training episodes re-rendered with WidowX arms, and the 334 real training episodes. Updates: the checkpoint used; Stage~3's single epoch has 26,811 updates, and we use its last saved checkpoint. Epochs count passes over the clips at an effective batch of 4 (Stage~2: 10,921 updates per epoch); the baseline epochs of \cref{tab:supp_training} count training episodes instead.}
\label{tab:supp_stages}
\end{table}

\paragraph{Stage 1.}
The conditioning canvas holds the RGB image, the sensor depth and a second copy of the RGB, and the history frame holds the same observation. The surface tile stores the depth $d$ of the nearest cloth surface, including cloth behind the robot, as $g(d) = 1 - (\log d - \log 0.2)/(\log 1.5 - \log 0.2)$ over 0.2--1.5\,m; a channel value $v > 0.1$ of the gap tile marks a hidden layer $(v - 0.1)/0.9 \times 0.15$\,m beneath the layer above it; and the mask tile is the visible-cloth mask. The stage is trained with a constant action input and the standard diffusion loss on all three tiles, on five-frame clips in which each frame is conditioned on its own observation; at inference, the $t_0$ observation fills the window and the first output frame of a single sample is kept. In the combination with the observation (Sec.~3.2), $k = \sum p\,o / \sum p^2$ over the pixels that take the observed depth, with $p$ the predicted and $o$ the observed surface depth; on the test episodes, at least 668 pixels enter this fit and $k$ lies between 0.926 and 1.025 (median 0.982). The point set has one point per pixel and layer wherever that layer has a depth; a hidden layer exists where its decoded gap is positive and the surface tile is not empty.

\paragraph{Stage 2.}
Per frame, the canvas has seven tiles (RGB, surface depth, gap tile and four flow tiles), so a 21-frame window is a $21\times4\times32\times224$ latent. The prior tiles store each axis of $\bar p_{i,t} - p_i$ as $\operatorname{clip}\big(\tfrac12 + (\bar p_{i,t} - p_i)/(0.6\,\mathrm{m}), 0, 1\big)$, and the residual tiles store
\begin{equation}
e_{i,t} = \operatorname{clip}\Big(\tfrac12 + \frac{r_{i,t}}{2\rho\,\sigma_t},\, 0,\, 1\Big), \qquad \rho = 1.5,
\label{eq:supp_target}
\end{equation}
with $r_{i,t}$ and $\sigma_t$ as in Sec.~3.3. The action tokens hold, per frame, the gripper position, grasp anchor and closed flag of both arms (14 values), embedded by a newly initialized three-layer MLP ($14\to1024\to1024\to1024$) of the Ctrl-World design~\cite{guo2026ctrl}; the gripper orientation enters only through the rigid prior. The six heat-map channels per frame are, for each arm, a Gaussian at the image projection of its gripper position, at every frame, and one at its grasp anchor while the grasp holds, together with the maximum of each type over the two arms. Each Gaussian has a standard deviation of 1.6 cells of the $32\times32$ grid (12.8 image pixels), and a point that projects outside the image gives an empty map. The channels are concatenated with the UNet input, whose first convolution is widened from 8 to 14 input channels; the weights of the six new channels start at zero, so the warm-started weights are unchanged at initialization.

The decoded-space terms of Sec.~3.3 are added to the latent diffusion loss $\mathcal{L}_\text{diff}$ as
\begin{equation}
\mathcal{L} = \mathcal{L}_\text{diff} + \sum_{j\in\{\text{flow},\,\text{anchor}\}} \gamma\,\frac{\bar{\mathcal{L}}_\text{diff}}{\bar{\mathcal{L}}_j}\,\min\big(\mathcal{L}_j,\, 3\bar{\mathcal{L}}_j\big),
\label{eq:supp_loss}
\end{equation}
with $\gamma = 0.5$ and $\bar{\cdot}$ an exponential moving average with decay 0.99; $\delta$ is measured in the encoded units of \cref{eq:supp_target}. Each term thus contributes at most 1.5 times the running diffusion loss; the cap bounds the value, not the gradient, so a step above it receives no gradient from that term.

\paragraph{Stage-2 checkpoint.}
The checkpoints after 14,000 and 16,000 updates were compared on four training episodes of the baselines' probe list (\cref{sec:supp_baselines}), one per class, by whole-episode rollouts of Stage~2 alone that start from the ground-truth point set at $t_0$ and use the same sampling seeds for both checkpoints. Their Chamfer distances to the ground truth are 15.2 and 15.5\,mm (between the predicted points and those of all labeled ground-truth layers, 8,000 of each, without post-processing, averaged over 1, 2, 3 and 4\,s and the last frame). The training loss, the full training objective averaged per update over each 2,000 updates, is 0.132, 0.117, 0.114, 0.110, 0.110, 0.105, 0.105 and 0.104 in the blocks ending at 2,000, 4,000, \dots, 16,000 updates; a line fitted to its means over 100 updates falls by 19\% per epoch from 8,000 to 12,000 updates (95\% interval 9 to 28\%) and has no significant trend over the last 4,000 updates before the checkpoint ($-3\%$ per epoch, $-14$ to $+7\%$; one epoch is 10,921 updates). With training at a plateau, we use the later, more trained checkpoint.

\paragraph{Settling.}
The table height is the median height of the observed $t_0$ depth in a 25-pixel ring around the predicted cloth mask, outside the robot mask. From the release frame on, the points are binned into an 8\,mm horizontal grid, and the $n$ points of a cell are stacked, in their height order at release, evenly over a height $h = (n/\bar n)\,h_0$ above the table, where $\bar n$ is the median number of points per occupied cell and $h_0$ the median height of the points above the table (at least 1.5\,mm), both at $t_0$. Each point keeps its horizontal position, and its height $z$ relaxes toward its target $z^\ast$ as $z \leftarrow z^\ast + (z - z^\ast)\,e^{-\Delta t/\tau}$ per frame, with $\Delta t = 0.1$\,s and $\tau = 0.15$\,s.

\paragraph{Depth rendering.}
The gap fill links each point of the frame-0 point set to its six nearest neighbors within 3\,cm. At every frame, each link longer than 6\,mm receives points at 1/3, 1/2 and 2/3 of its length, which are kept only where they land on pixels that no projected point covers. All points are then projected into a $256\times256$ depth image, keeping the nearest depth per pixel. The depth repair (i)~keeps the main cloth region: the occupancy, box-filtered over $5\times5$ pixels, is thresholded at 0.3 and split into connected components, of which the largest is kept, together with the second largest if it reaches 20\% of the largest, including the holes they enclose; (ii)~replaces spikes, in up to four passes, by giving each pixel that deviates by more than 3\,cm from the median of its $11\times11$ neighborhood the mean of the non-spike pixels in its $9\times9$ neighborhood; and (iii)~gives the remaining holes inside the kept region the nearest valid depth.

\paragraph{Stage 3.}
The conditioning tile is the predicted depth of the cloth surface alone, encoded with $g$ as in Stage~1, with the robot mask painted in magenta. In training, the tiles use the ground-truth cloth-surface depth and the robot masks stored with the benchmark, which leave out robot pixels behind cloth; at inference, the robot mask of Sec.~3.1 is painted. The action encoder receives a constant.

\section{Metrics}
\label{sec:supp_metrics}

\paragraph{Geometry.}
ClothCast and the baselines are scored with the same pixel selection and Chamfer computation. In each predicted frame, the evaluation pixels are the ground-truth visible-cloth pixels outside the robot mask, eroded twice with a $3\times3$ kernel and restricted to valid ground-truth depth, and both depth maps are back-projected at these pixels into camera coordinates. Flex-$\pi$ and DepthWorld predict geometry at every pixel. A pixel at which ClothCast predicts no depth has no predicted point, while its ground-truth point is matched to the nearest predicted point elsewhere. The 6 of 29,759 predicted frames in which ClothCast covers no evaluation pixel have no Chamfer distance and count only in the coverage.

\paragraph{RGB.}
All methods are scored with the same implementation of PSNR, SSIM and LPIPS (VGG)~\cite{zhang2018lpips}. Cloth PSNR excludes the robot pixels; cloth SSIM is averaged over the cloth region, and its windows reach robot pixels only at the region boundary, which gives them 0.2\% of the window weight on average; cloth LPIPS is computed on a box around the cloth region, which is on average 22\% robot and 23\% cloth. Pasting the true arm into ClothCast's predicted frames lowers its cloth LPIPS by 0.005 and raises its cloth SSIM by 0.003.

\paragraph{Aggregation.}
Every metric is averaged over the predicted frames of an episode, excluding $t_0$, then over the episodes of a class, then over the four classes. The horizons of 1 to 4\,s are the 10th to 40th frames after $t_0$, and Last is each episode's final frame.

\section{Baselines}
\label{sec:supp_baselines}

\paragraph{Inputs.}
Each baseline starts from the observation at $t_0$ (RGB-D for Flex-$\pi$ and DepthWorld, RGB for Cosmos-Predict2.5 and V-JEPA~2-AC) and receives the planned trajectory as the benchmark's 14-D actions, per-arm increments of gripper translation, rotation and closure; in simulation these equal the differences of the recorded gripper states, which under kinematic replay equal the planned ones (largest deviation $3.7\times10^{-9}$\,m). Flex-$\pi$ also receives the initial gripper state and the camera intrinsics, and V-JEPA~2-AC the gripper state at every frame. No baseline receives the camera pose or the robot model. Each rolls out the whole episode autoregressively from its own predictions.

\paragraph{Training.}
\Cref{tab:supp_training} lists how each baseline is adapted. All start from their released checkpoints and train on the 9,484 training episodes.

\begin{table*}[t]
\centering
\footnotesize
\setlength{\tabcolsep}{4pt}
\newcommand{\twol}[2]{\begin{tabular}[t]{@{}l@{}}#1\\#2\end{tabular}}
\begin{tabular}{@{}lllll@{}}
\toprule
Method & Backbone & Adapted & LR & Tested: updates (epochs) \\
\midrule
DepthWorld~\cite{depthworld} & \twol{SVD UNet (Ctrl-World design),}{DPT point-map head} & \twol{UNet, action MLP,}{DPT head; EMA 0.999} & $1{\times}10^{-5}$ & EMA weights at 6,371 (2.69) \\
\addlinespace
Flex-$\pi$~\cite{flexpi} & \twol{Wan2.2-TI2V DiT (5B),}{action DiT; frozen} & LoRA, 14-D action heads & $5{\times}10^{-5}$ & 8,100 (3.42) \\
\addlinespace
Cosmos-Predict2.5~\cite{cosmos} & \twol{action-conditioned model}{(2B); frozen} & \twol{LoRA (rank 16),}{action embedders} & $1{\times}10^{-4}$ & 4,871 (2.05) \\
\addlinespace
V-JEPA~2-AC~\cite{vjepa2ac} & \twol{ViT-g encoder (1B), frozen;}{predictor (300M)} & predictor & $1{\times}10^{-5}$ & 14,022 at batch 2 (2.96) \\
 & & new RGB decoder & $2{\times}10^{-4}$ & 23,710 at batch 16 (10) \\
\bottomrule
\end{tabular}
\caption{\textbf{Baseline training.} All baselines start from their released checkpoints and train on the 9,484 training episodes, with an effective batch of 4 unless noted, so that one epoch, a pass over the 9,484 episodes, has 2,371 updates (4,742 for the V-JEPA~2-AC predictor at batch 2; for the V-JEPA~2-AC RGB decoder, at batch 16, an epoch is likewise counted as 2,371 updates); LR: learning rate. DepthWorld's DPT head predicts its per-pixel geometry, listed as per-pixel depth in Tab.~1, as a point map; its objective is its diffusion loss plus 0.05 times its point-map loss, with AdamW, weight decay 0.01 and gradient clipping at 1.0. Every test number of a baseline comes from its tested checkpoint.}
\label{tab:supp_training}
\end{table*}

\paragraph{Probe and stopping.}
Each baseline's checkpoints are scored on a fixed, class-balanced probe of 20 training episodes: whole-episode rollouts on 4 of them (one per class) and, on the other 16 (four per class), rollouts of one fixed 21-frame window from the middle of the episode, starting from the ground-truth observation at the window's first frame. The probe score averages the class-balanced whole-episode and window scores, measured by the cloth Chamfer distance of Sec.~5.1 for Flex-$\pi$ and DepthWorld and by cloth LPIPS for Cosmos and the V-JEPA~2-AC predictor; the V-JEPA~2-AC RGB decoder is scored by its class-balanced L1 reconstruction error on 64 fixed training frames (16 episodes, four frames each). Scoring runs every 500 updates, from the end of the first epoch for Flex-$\pi$, Cosmos and DepthWorld. Training was stopped manually, by inspecting these scores, once they had leveled off (Sec.~5.1), except for the V-JEPA~2-AC decoder, which trained for the full 10 epochs; no test episode is read for stopping or checkpoint selection. \Cref{tab:supp_probe_a,tab:supp_probe_b} give every probe score, the evidence behind each stop. The scores fluctuate from probe to probe, and no stopped run has a significant trend before its stop: a line fitted to its last eight probe scores (Cosmos: its five complete ones) changes the score per epoch by $-4\%$ for Flex-$\pi$ (95\% interval $-24$ to $+16\%$), $+3\%$ for DepthWorld ($-18$ to $+24\%$), $-6\%$ for Cosmos ($-17$ to $+6\%$) and $-4\%$ for the V-JEPA~2-AC predictor ($-12$ to $+4\%$), relative to the mean of the fitted scores. Each run is tested at its last probed or last saved checkpoint, not at a checkpoint chosen by its probe score: DepthWorld was stopped after 6,597 updates, and its EMA weights at the last probed checkpoint, after 6,371 updates, were tested.

\begin{table*}[t]
\centering
\footnotesize
\setlength{\tabcolsep}{3.2pt}
\begin{tabular}{@{}lcccccccccccc@{}}
\toprule
Updates & 2,371 & 2,871 & 3,371 & 3,871 & 4,371 & 4,871 & 5,371 & 5,871 & 6,371 & 6,871 & 7,371 & 7,871 \\
Epochs & 1.00 & 1.21 & 1.42 & 1.63 & 1.84 & 2.05 & 2.27 & 2.48 & 2.69 & 2.90 & 3.11 & 3.32 \\
\midrule
Flex-$\pi$ (CD, mm) & 60.50 & 38.60 & 45.20 & 47.02 & 39.19 & 39.59 & 43.67 & 38.36 & 43.16 & 40.18 & 30.50 & 42.41 \\
DepthWorld (CD, mm) & 58.31 & 59.71 & 51.50 & 49.54 & 58.44 & 66.07 & 59.37 & 48.48 & \textbf{60.54} &  &  &  \\
Cosmos (LPIPS) & 0.107 & 0.101 & 0.105 & 0.099 & 0.101 & -- &  &  &  &  &  &  \\
\bottomrule
\end{tabular}
\caption{\textbf{Probe scores} of Flex-$\pi$, DepthWorld and Cosmos every 500 updates from the end of the first epoch (lower is better; bold: the tested checkpoint). Flex-$\pi$ was tested at 8,100 updates, which were not probed, and Cosmos at 4,871 updates, whose probe did not complete (--).}
\label{tab:supp_probe_a}
\end{table*}

\begin{table*}[t]
\centering
\footnotesize
\setlength{\tabcolsep}{3.6pt}
\begin{tabular}{@{}lcccccccccc@{}}
\toprule
\multicolumn{11}{@{}l}{\emph{V-JEPA~2-AC predictor}: cloth LPIPS (one epoch = 4,742 updates at batch 2)} \\
\midrule
Updates & 22 & 522 & 1,022 & 1,522 & 2,022 & 2,522 & 3,022 & 3,522 & 4,022 & 4,522 \\
LPIPS & 0.145 & 0.144 & 0.144 & 0.139 & 0.137 & 0.137 & 0.142 & 0.135 & 0.137 & 0.140 \\
\midrule
Updates & 5,022 & 5,522 & 6,022 & 6,522 & 7,022 & 7,522 & 8,022 & 8,522 & 9,022 & 9,522 \\
 & 0.133 & 0.130 & 0.131 & 0.125 & 0.129 & 0.125 & 0.125 & 0.138 & 0.128 & 0.129 \\
\midrule
Updates & 10,022 & 10,522 & 11,022 & 11,522 & 12,022 & 12,522 & 13,022 & 13,522 & 14,022 &  \\
 & 0.128 & 0.132 & 0.125 & 0.126 & 0.125 & 0.127 & 0.131 & 0.126 & \textbf{0.123} &  \\
\midrule
\multicolumn{11}{@{}l}{\emph{V-JEPA~2-AC RGB decoder}: reconstruction L1 $\times 100$ (one epoch = 2,371 updates at batch 16)} \\
\midrule
Updates & 0 & 500 & 1,000 & 1,500 & 2,000 & 2,500 & 3,000 & 3,500 & 4,000 & 4,500 \\
L1 & 11.22 & 2.39 & 2.01 & 1.82 & 1.75 & 1.65 & 1.53 & 1.49 & 1.46 & 1.45 \\
\midrule
Updates & 5,000 & 5,500 & 6,000 & 6,500 & 7,000 & 7,500 & 8,000 & 8,500 & 9,000 & 9,500 \\
 & 1.45 & 1.36 & 1.32 & 1.32 & 1.28 & 1.30 & 1.26 & 1.25 & 1.23 & 1.25 \\
\midrule
Updates & 10,000 & 10,500 & 11,000 & 11,500 & 12,000 & 12,500 & 13,000 & 13,500 & 14,000 & 14,500 \\
 & 1.22 & 1.22 & 1.16 & 1.19 & 1.19 & 1.16 & 1.14 & 1.14 & 1.12 & 1.13 \\
\midrule
Updates & 15,000 & 15,500 & 16,000 & 16,500 & 17,000 & 17,500 & 18,000 & 18,500 & 19,000 & 19,500 \\
 & 1.10 & 1.12 & 1.09 & 1.11 & 1.09 & 1.08 & 1.07 & 1.07 & 1.08 & 1.05 \\
\midrule
Updates & 20,000 & 20,500 & 21,000 & 21,500 & 22,000 & 22,500 & 23,000 & 23,500 & 23,710 &  \\
 & 1.08 & 1.05 & 1.06 & 1.04 & 1.04 & 1.06 & 1.02 & 1.05 & \textbf{1.06} &  \\
\bottomrule
\end{tabular}
\caption{\textbf{Probe scores of V-JEPA~2-AC} every 500 updates (lower is better; bold: the tested checkpoint).}
\label{tab:supp_probe_b}
\end{table*}

\section{Per-horizon results}
\label{sec:supp_horizon}

\Cref{tab:supp_rgb_horizon} gives the cloth PSNR of Tab.~3 over the prediction horizon, and \cref{tab:supp_ablation_horizon} the Chamfer distance and coverage of every variant of Tab.~5.

\begin{table}[t]
\centering
\small
\setlength{\tabcolsep}{4.5pt}
\begin{tabular}{@{}lcccccc@{}}
\toprule
Method & 0.1\,s & 1\,s & 2\,s & 3\,s & 4\,s & Last \\
\midrule
DepthWorld & 29.1 & 22.0 & 13.8 & 11.6 & 11.7 & 11.3 \\
Flex-$\pi$ & \textbf{31.1} & 19.3 & 13.7 & 12.5 & 13.8 & 12.7 \\
V-JEPA~2-AC & 26.4 & 23.9 & 18.0 & 14.6 & 14.4 & 13.2 \\
Cosmos & 31.0 & 23.0 & 16.7 & 14.4 & 14.5 & 13.2 \\
ClothCast & 28.2 & \textbf{24.4} & \textbf{20.5} & \textbf{19.5} & \textbf{19.2} & \textbf{18.1} \\
\bottomrule
\end{tabular}
\caption{\textbf{Cloth PSNR (dB) over the prediction horizon} on the 492 test episodes; 0.1\,s is the first predicted frame.}
\label{tab:supp_rgb_horizon}
\end{table}

\begin{table}[t]
\centering
\small
\setlength{\tabcolsep}{3.0pt}
\begin{tabular}{@{}lccccccc@{}}
\toprule
& \multicolumn{6}{c}{Chamfer distance (mm) $\downarrow$} & \\
\cmidrule(lr){2-7}
Variant & 1\,s & 2\,s & 3\,s & 4\,s & Last & All & Cov. \\
\midrule
ClothCast & 7.6 & 15.7 & 19.1 & 17.9 & 15.8 & 14.5 & 0.90 \\
\quad w/o settling & 7.6 & 15.7 & 19.1 & 20.1 & 20.8 & 15.5 & 0.90 \\
\quad w/o depth repair & 7.8 & 15.9 & 19.5 & 18.1 & 15.6 & 14.7 & 0.87 \\
\quad raw points & 7.7 & 15.9 & 19.5 & 20.6 & 21.1 & 15.8 & 0.76 \\
\midrule
Rigid prior only & 15.9 & 31.1 & 28.2 & 22.0 & 15.9 & 21.2 & 0.81 \\
\quad raw points & 15.7 & 29.9 & 27.7 & 35.0 & 36.5 & 24.8 & 0.72 \\
\midrule
Camera from $t_0$ & 7.5 & 15.8 & 19.4 & 17.4 & 15.6 & 14.3 & 0.90 \\
\bottomrule
\end{tabular}
\caption{\textbf{Ablations over the prediction horizon} on the 50 ablation episodes (Sec.~5.4). \emph{Rigid prior only}: the learned residual set to zero, with the prior of Eq.~(1) chained over the same windows and the same gap fill and post-processing. \emph{w/o}: the named post-processing step removed; \emph{raw points}: the points projected directly, without gap fill or post-processing. \emph{Camera from $t_0$}: pose recovered from the $t_0$ frame. Cov.: coverage over all frames.}
\label{tab:supp_ablation_horizon}
\end{table}

\section{Real-robot evaluation}
\label{sec:supp_real}

\paragraph{Data.}
The 418 real episodes show two WidowX~AI arms carrying, dragging, lifting, pulling apart or swinging seven items: two gingham cloths, a towel, a polo shirt, a t-shirt and two pillow covers. Each episode provides $256\times256$ RGB-D frames at 10\,Hz from a RealSense D455 camera, the camera calibration, the measured gripper poses and jaw states, and approximate masks of the visible cloth (grown from color) and of the robot (rendered from the measured joints, without the pixels where the sensor depth is at least 2\,cm closer than the arm); there is no cloth mesh, so the hidden layers have no ground truth. The forecast starts at the frame at which the grasp is established. The data has no predefined split, so we draw 84 test episodes at random and keep the other 334 for training; the same seven items appear in both.

\paragraph{Adaptation and inputs.}
Stage~2 is used as trained in simulation. Stage~1 is fine-tuned from its simulation weights for one epoch on simulated training episodes re-rendered with WidowX arms (\cref{tab:supp_stages}), so neither geometry stage is trained on a real image. Stage~3 is fine-tuned from its simulation weights for three epochs on clips of the real training episodes, conditioned on the sensor depth of the visible cloth, with sensor holes filled by the nearest valid depth and the dataset's robot mask painted in, and supervised with the real images; at test time, it receives the predicted cloth depth and, at future frames, robot masks rendered from the joints with the WidowX meshes alone, without the sensor depth. ClothCast receives the measured gripper trajectory, normalized and clipped with the same bounds as in simulation, and settling uses a fixed $h_0 = 20$\,mm (Sec.~5.5).

\paragraph{Metrics.}
The geometry is scored with the Chamfer distance and coverage of Sec.~5.1 against the D455 depth, on evaluation pixels taken from the dataset's approximate visible-cloth and robot masks; frames are averaged per episode, episodes per item, and then over the seven items. The sensor depth is noisy and the masks are approximate, so these errors are not directly comparable with the simulated ones.

\paragraph{Results.}
\Cref{tab:supp_real} gives the geometry per horizon, for ClothCast and for the rigid prior alone with the same Stage-1 points and post-processing, and \cref{tab:supp_real_items} per item. Averaged per item, as in Sec.~5.5, the learned residual lowers the error over all frames by 9.6\,mm, from 44.9 to 35.3\,mm; it helps at 1 to 3\,s, matches the prior at 4\,s (41.6 vs.\ 41.8\,mm) and is worse at 0.1\,s and at the last frame, and on the pillow covers also from 3\,s on. Paired per episode, with every episode weighted equally, the rigid prior's error over all frames is 10.2\,mm higher than ClothCast's (95\% interval 7.3 to 13.3\,mm), and ClothCast's is lower in 70 of the 84 episodes. Part of the pillow covers' higher error comes from the motions: 16 of their 20 test episodes are carries, the motion with the highest error for every item, against 32 of the other 64. Within carries, the pillow-cover episodes average 54.2\,mm against 36.1\,mm for the other items; in the four pillow-cover episodes of other motions, 25.6\,mm against 25.4\,mm. Over all 84 episodes, the cloth PSNR is 14.6\,dB, the cloth SSIM 0.17 and the cloth LPIPS 0.137; the cloth PSNR is 15.3, 14.3, 13.9, 13.4 and 13.2\,dB at 1, 2, 3 and 4\,s and at the last frame. We evaluate ClothCast only; no baseline was adapted to the real robot.

\begin{table*}[t]
\centering
\footnotesize
\setlength{\tabcolsep}{4.5pt}
\begin{tabular}{@{}llrcccccccc@{}}
\toprule
& & & \multicolumn{7}{c}{Chamfer distance (mm) $\downarrow$} & \\
\cmidrule(lr){4-10}
Test episodes & Method & $n$ & 0.1\,s & 1\,s & 2\,s & 3\,s & 4\,s & Last & All & Cov. \\
\midrule
Cloths, towel, shirts & ClothCast & 64 & 5.7 & 27.7 & 35.2 & 36.9 & 35.9 & 27.7 & 30.2 & 0.69 \\
 & Rigid prior & 64 & 5.1 & 50.9 & 60.7 & 48.8 & 36.8 & 24.5 & 41.7 & 0.70 \\
\midrule
Pillow covers & ClothCast & 20 & 5.2 & 44.8 & 48.2 & 62.5 & 55.9 & 52.3 & 48.1 & 0.60 \\
 & Rigid prior & 20 & 4.9 & 63.4 & 69.1 & 54.4 & 54.3 & 46.5 & 52.9 & 0.59 \\
\midrule
All & ClothCast & 84 & 5.6 & 32.6 & 38.9 & 44.2 & 41.6 & 34.7 & 35.3 & 0.66 \\
 & Rigid prior & 84 & 5.0 & 54.5 & 63.1 & 50.4 & 41.8 & 30.8 & 44.9 & 0.67 \\
\bottomrule
\end{tabular}
\caption{\textbf{Real-robot geometry}: visible-surface Chamfer distance against the D455 depth and coverage over all frames, for ClothCast and for the rigid prior alone with the same Stage-1 points and post-processing. All 84 episodes reach 1\,s; 83, 82 and 66 reach 2, 3 and 4\,s (pillow covers: 19, 19 and 13).}
\label{tab:supp_real}

\vspace{0.8em}
\centering
\footnotesize
\setlength{\tabcolsep}{6pt}
\newcommand{\twoc}[2]{\begin{tabular}[t]{@{}c@{}}#1\\#2\end{tabular}}
\begin{tabular}{@{}lccccccc@{}}
\toprule
& \twoc{Teal}{t-shirt} & \twoc{Black gingham}{cloth} & \twoc{Striped}{polo shirt} & \twoc{Red gingham}{cloth} & \twoc{Red grid}{towel} & \twoc{Teal paisley}{pillow cover} & \twoc{Floral}{pillow cover} \\
\midrule
Chamfer distance (mm) & 25.0 & 29.1 & 31.2 & 31.5 & 34.1 & 43.7 & 52.4 \\
Coverage & 0.76 & 0.68 & 0.68 & 0.66 & 0.65 & 0.59 & 0.61 \\
\bottomrule
\end{tabular}
\caption{\textbf{Real-robot geometry per item}: ClothCast's visible-surface Chamfer distance and coverage over all frames.}
\label{tab:supp_real_items}
\end{table*}

\clearpage
{
    \small
    \bibliographystyle{ieeenat_fullname}
    \bibliography{main}
}

\end{document}